%% file: main.tex
\pdfoutput=1 
\newif\ifjds
\jdsfalse    

\ifjds
  \documentclass[linenumber]{jdsart}
\else
  \documentclass[11pt]{article}
\fi

\usepackage{amsmath,amssymb,bm}
\usepackage{graphicx}
\usepackage{xcolor}
\usepackage{booktabs}
\usepackage{siunitx}
\usepackage{algorithm}
\usepackage{algpseudocode}
\usepackage{subcaption}
\usepackage{tikz}
\usetikzlibrary{positioning,calc}

\ifjds
\else
  \usepackage[margin=1in]{geometry}
  \usepackage{amsthm}
  \usepackage[round]{natbib}   
\fi

\input{bold_letters}

\ifjds\else
  \usepackage[colorlinks=true,citecolor=blue,linkcolor=blue,urlcolor=blue]{hyperref}
\fi

\begin{document}

\ifjds
\else

\title{Adaptive Conformal Prediction for Image Regression Models with Application to an Inertial Confinement Fusion Emulator}
\author{
Carrie J. Lei-Cramer\thanks{Department of Statistics, Texas A\&M University, College Station, TX, USA.}
\and
Michael S. Jones\thanks{Lawrence Livermore National Laboratory, Livermore, CA, USA.}
\and
Laura J. Wendelberger\thanks{Lawrence Livermore National Laboratory, Livermore, CA, USA. Corresponding author: \texttt{wendelberger1@llnl.gov}}
}
\date{}
\maketitle

\input{abstract}
\fi

\input{introduction}

\input{motivation}
\input{background}

\input{method}

\input{application}

\input{conclusion}

\ifjds
\else

\section*{Acknowledgments}
This work was performed under the auspices of the U.S. Department of Energy by Lawrence Livermore National Laboratory under Contract DE-AC52-07NA27344. Release number: LLNL-MI-2013347.

\bibliographystyle{apalike}
\bibliography{references}

\newpage
\appendix

\input{appendix}
\fi

\end{document}

%% file: bold_letters.tex
\newcommand{\bx}{\mathbf{x}}



%% file: abstract.tex
\begin{abstract}

Uncertainty quantification is critical in scientific machine learning, where black-box, image-based models are increasingly deployed in high-stakes settings. In many such applications, model outputs inform costly decisions, yet most methods provide only point estimates without quantifying predictive uncertainty. This challenge is compounded by the limited accessibility and interpretability of model internals, making it difficult to assess reliability across different regions of the input space. As a result, there is a growing need for methods that can provide input-dependent uncertainty estimates to guide both model development and downstream experimentation.
To address this need, we propose Adaptive Conformal Prediction using Nearest Neighbors (ACPNN), an input-adaptive conformal framework for image regression. ACPNN leverages information from neighboring samples to produce locally adaptive uncertainty estimates while maintaining low computational cost. The neighborhood structure is defined using a scaled distance metric learned via a Gaussian Process with an automatic relevance determination (ARD) kernel. We demonstrate the effectiveness of ACPNN on a diffusion model for emulating inertial confinement fusion (ICF) simulations, showing that it achieves reliable and adaptive uncertainty quantification.

\end{abstract}

%% file: introduction.tex
\section{Introduction}
Uncertainty quantification (UQ) in machine learning (ML) has become increasingly important as ML models are more widely adopted in scientific research, particularly in high-stakes domains such as inertial confinement fusion (ICF) (\cite{BenTayeb2024}, \cite{Anirudh9741}, \cite{Spears2025}), where both simulations and experiments are costly. ICF seeks to induce fusion by confining matter under extreme pressure and temperature and has the potential to produce clean energy at scale; however, its success is highly sensitive to experimental design under extreme pressure and temperature. High-fidelity multiphysics simulations can model the forward problem, but they are computationally expensive. To address the inverse problem—identifying input designs that achieve desired outputs—ML-based emulators (e.g., \cite{kang2022self}) have been developed. Despite their efficiency, these models often lack rigorous uncertainty quantification, raising critical concerns about their reliability as “black-box” predictors.
These challenges are further amplified when working with image-based data, which introduces additional complexity compared to scalar quantities. Incorporating UQ improves confidence in model predictions, which helps guide experimental design across a range of fields—such as protein folding (\cite{Jumper2021}, \cite{Baek2021}) and material discovery (\cite{Butler2018}, \cite{Gao2023})—where confidence can vary dramatically across predicted structures.

A wide range of methods have been developed to achieve UQ in the field of computer vision, including Deep Ensembles \citep{lakshminarayanan2017simple,kallnielsen2021deep} and Bayesian Neural Networks \citep{kendall2017what,blundell2015weight}, which often works with approximation techniques like Monte Carlo Dropout \citep{lemay2022improving,abdollahi2023applying, gal2016dropout}. Bayesian methods depend on prior distributions that may be misspecified and are usually computationally expensive, while ensemble-based methods provide heuristic uncertainty estimates. Moreover, these approaches requires significant modification to the model architecture and typically lack formal guarantees of reliability. 

The conformal prediction (CP) framework has attracted substantial attention as a  model-agnostic approach to uncertainty quantification. Its key advantage lies in providing coverage guarantee under minimal assumptions, i.e. its prediction sets contain the true outcome with a user-specified significance level, regardless of the underlying model. It requires only one key assumption: data exchangeability, i.e. the joint distribution of data is permutation invariant. This condition can be implied by independent and identically distributed (i.i.d.), which is a common assumption used in statistics. 
Beyond its coverage guarantees, the potential input-adaptiveness of CP further enhances its utility in scientific discovery, as it can help identify regions of high uncertainty, thereby guiding model refinement and informing future experimental directions.

CP has been incorporated into various computer vision models \citep{angelopoulos2021uncertainty, roy2025are, degrancey2022object}, including input-adaptive approaches for tasks like image segmentation \citep{bereska2025sacp, luo2026enhancing} and object detection \citep{timans2024conformalod}. 
Although the related works also involve image-input models, they primarily analyze pixel-level uncertainty, while our method concentrates on the image's overall structure.

In this work, we propose Adaptive Conformal Prediction with Nearest Neighbors (ACPNN), a novel conformal prediction approach designed to produce input-adaptive conformal sets using information from nearest neighbors. Building on the Adaptive Conformal Prediction framework of \cite{Papadopoulos2011}, our method extends adaptive uncertainty quantification to high-dimensional input spaces, such as images, where classical approaches may struggle to capture local structure. Our method fills a gap in adaptive UQ by producing statistically reliable prediction sets for image regression models. We apply ACPNN to an ICF dataset and evaluate several CP approaches.

The remainder of this paper is organized as follows. Section \ref{sec:motivation} presents the application motivating our method. Section \ref{sec:methodology} reviews building blocks of our proposed method and introduces our proposed method. Section \ref{sec:application} demonstrates the validity and adaptiveness of our proposed method, benchmarked with other methods.

%% file: motivation.tex
\section{Motivation: Inertial Confinement Fusion}\label{sec:motivation}

Inertial confinement fusion (ICF) aims to produce energy by compressing and heating hydrogen isotopes to extreme temperatures and pressures, offering a potential path toward clean and sustainable energy production \citep{Nuckolls1972}. In 2022, controlled ignition at the National Ignition Facility (NIF) achieved the first positive net energy gain \citep{Abu-Shawareb2024}, \citep{Kritcher2024}.

Achieving ignition and high fusion gain requires precise control of energy delivery and capsule symmetry, as well as extremely high temperatures and pressures. Because these experiments are resource‑intensive and limited in number, researchers often rely on large‑scale radiation‑hydrodynamics simulations such as HYDRA \citep{Marinak_2001} to explore capsule designs and predict experimental outcomes. These physics‑based simulators generate datasets comprising scalar quantities and images that capture implosion dynamics.

\begin{figure}
	\centering
	\includegraphics[trim={0cm, 5cm, 0cm, 5cm},clip, width =.98\linewidth]{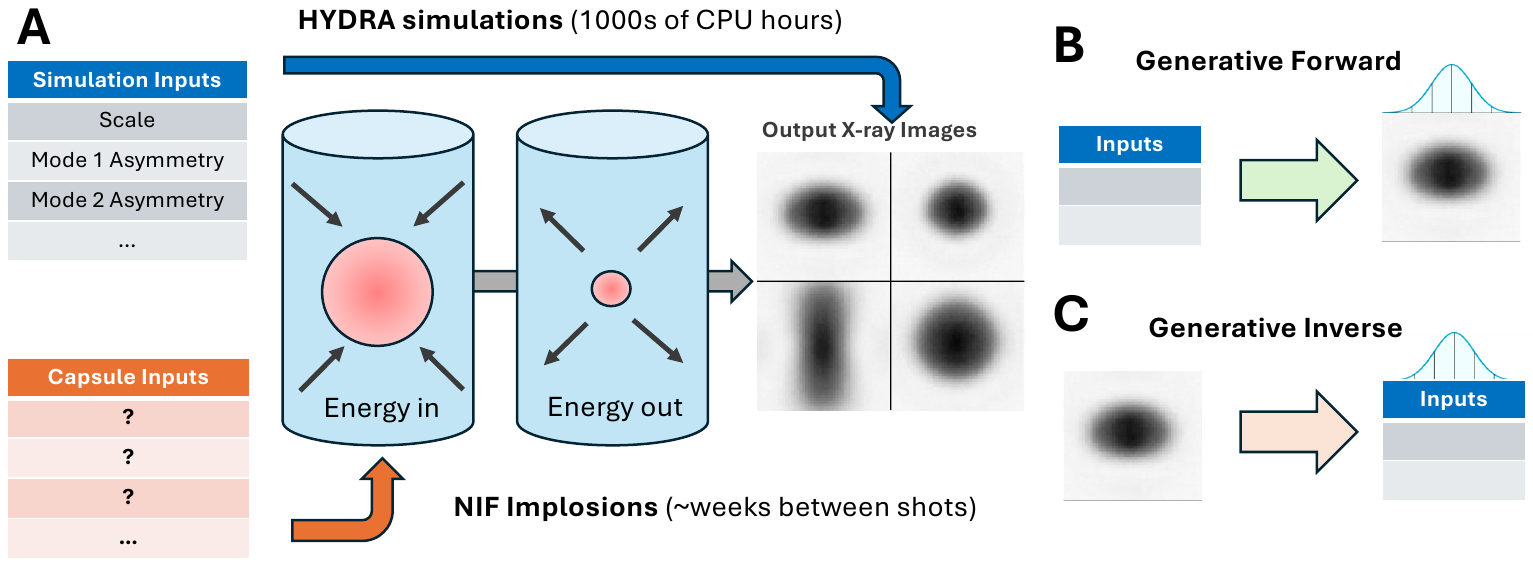}
    \caption{Schematic overview of ICF simulation, experiment, and emulation workflows. 
    (A) HYDRA simulations, driven by input parameters such as capsule scale and asymmetry, are performed in parallel with NIF implosion experiments. Both workflows produce diagnostic X-ray images; however, HYDRA simulations require substantial computational resources, while NIF experiments are limited by the weeks-long timescale between shots.
    (B) In the generative forward setting, models are conditioned on input parameters to produce corresponding diagnostic X-ray images.
    (C) In the generative inverse setting, models are conditioned on diagnostic X-ray images to produce input parameter values.
}
    \label{fig:model}
\end{figure}

Recent advances in machine learning enables emulation of the fusion process at lower computational cost, for example the diffusion‑based foundation models by \cite{jones2026} and autoencoder‑based neural‑network models by \cite{Anirudh9741}. The former supports image‑only inverse prediction (see illustration in Figure \ref{fig:model})—i.e., inferring capsule inputs directly from X-ray and neutron images—which is valuable for diagnosing degradations to experiments and improving future designs.

Despite these advances, such models often lack reliable uncertainty quantification. Without it, assessing a model’s ability to capture underlying relationships is difficult, and it becomes hard to determine which regions of the input space are physically reliable or experimentally consistent. Establishing methods to adaptively quantify uncertainty across the input space is therefore critical for emulator validation, interpretation of experimental data, and, ultimately, for guiding the design of next‑generation ICF targets with reproducible, high‑yield performance.

We aim to generate uncertainty-quantifying prediction sets for a \textquotedblleft black-box" model that predicts numeric values from images, ensuring these sets maintain a targeted average coverage rate. What's more, the bandwidth of these sets are input adaptive, reflecting uncertainty of model prediction at different input regions. Formally, given a trained model $f: \mathbb{R}^d \to \mathbb{R}$, a new image $\bx \in \mathbb{R}^d$, and a confidence level $\alpha$, we seek to produce an adaptive prediction set $\mathcal{C}(\bx)$ such that $P\!\big(y \in \mathcal{C}(\bx)\big) \ge 1 - \alpha$.

%% file: background.tex
\section{Methodology} \label{sec:methodology}

\subsection{Conformal Prediction} \label{sec:background_cp}
Introduced by \cite{saunders1999}, conformal prediction (CP) is a framework that provides statistically reliable uncertainty quantification for machine learning models by constructing prediction sets or intervals with guaranteed coverage performance.
Unlike traditional point predictions, conformal methods output sets of possible labels (in classification) or intervals (in regression) that contain the true outcome with a user-specified confidence level 1-$\alpha$, such as 90\% or 95\%. The core idea is to measure how \textquotedblleft conformal", or consistent, a new data point is with previously observed data using a nonconformity score, and then calibrate thresholds through permutation - or exchangeability-based arguments. Crucially, permuting the random variables does not change their joint distribution, and therefore the rank of the new data point among all data points is uniformly distributed from 1 to the total number of variables. 
Since CP relies on empirical quantiles, its guarantees are agnostic to model correctness and the form of the error distribution, provided the data are exchangeable.
In other words, conformal prediction is distribution-free, model-agnostic, and computationally simple to apply on top of any black-box predictor, offering a practical and theoretically sound way to quantify prediction uncertainty. 

\textbf{Conformal prediction:} In the split conformal prediction framework \citep{vovk2005algorithmic}, the dataset
\(\mathcal{D} = \{Z_i =(\mathbf{x}_i, y_i)\}_{i=1}^n\), 
where \(\mathbf{x}_i \in \mathcal{X} \subseteq \mathbb{R}^d\) 
and \(y_i \in \mathcal{Y}\subseteq \mathbb{R} \) is randomly spitted into two disjoint subsets $\mathcal{D}_{\text{train}}$ and $\mathcal{D}_{\text{cal}}$. These two subsets are defined as \(\mathcal{D}_{\text{train}} = \{(\mathbf{x}_i, y_i) : i \in I_{\text{train}}\}\) 
and 
\(\mathcal{D}_{\text{cal}} = \{(\mathbf{x}_i, y_i) : i \in I_{\text{cal}}\}\), where \(I_{\text{train}}\) and \(I_{\text{cal}}\) are subsets of indices from 1 to $n$,
 such that 
\(I_{\text{train}} \cap I_{\text{cal}} = \emptyset\), 
\(I_{\text{train}} \cup I_{\text{cal}} = \{1, \ldots, n\}\). Denote $n_1 = |I_{\text{train}}|, n_2 = |I_{\text{cal}}|, n = n_1 + n_2 $.

Given a predicted model $f$ that is trained on $\mathcal{D}_{train}$, and a new testing data point $Z_{n+1} = (\mathbf{x}_{n+1}, y_{n+1}) \in \mathcal{D}_{test}$ and $\mathcal{D}\cap \mathcal{D}_{test} = \emptyset$, conformal prediction uses absolute error as a (non)conformity score: $s_i = |f(\bx_i) - y_i|, i\in \mathcal{I}_{cal}$ to form prediction intervals. 
With the assumption that $\mathcal{D}_{cal} \cup \{Z_{n+1}\} $ are exchangeable,
the $(1-\alpha)$ quantile of scores $q_{1-\alpha}$ determines the conformal interval: $\mathcal{C}(\bx_{n+1}) = [{f}(\bx_{n+1}) - q_{1-\alpha}, {f}(\bx_{n+1}) + q_{1-\alpha}]$. This ensures finite-sample marginal coverage at level $(1-\alpha)$, such that: 
$$\mathbb{P}\big( y_{n+1} \in \mathcal{C}_{1-\alpha}(\bx_{n+1}) \big) \geq 1-\alpha.$$ 
The method for applying conformal prediction is shown in Algorithm \ref{alg:vcp}.

\textbf{Scaled conformal prediction: }Conformal intervals constructed produced by conformal prediction produced intervals of equal widths, regardless of the input values. Scaled conformal prediction is a variation of conformal prediction designed to improve the efficiency of prediction intervals, particularly in regression tasks where the magnitude of uncertainty varies across inputs (heteroskedasticity). Instead of using raw residuals from a fitted model as nonconformity scores, scaled conformal prediction divides each residual by an estimate of the local variability $\hat\sigma(\mathbf{\bx}_i)$, which is a measure of the dispersion of the residuals at  $\mathbf{\bx}_i$. Common choices of  $\mathbf{\bx}_i$ are conditional mean absolute deviation (MAD) of $\big|Y_i - \bar{f}(\bx) \big|$ given $X = \bx $ \citep{romano2019} or standard deviation of  $f(x)$ \citep{angelopoulos2022}. The conformal scores then take the mathematical form: $s_i = \frac{|\bar{{f}}(\bx_i) - y_i|}{std({f}(\bx_i))}$. This scaling step adjusts the widths of prediction intervals so that they expand in regions with high uncertainty and shrink in regions with low uncertainty, leading to intervals that are more adaptive and informative while still maintaining the rigorous marginal coverage guarantees of conformal prediction.

\textbf{Conformal prediction with nearest neighbors: } \cite{Papadopoulos2011} proposed new nonconformal scores to make the measure more adaptive to local inputs by leveraging information from nearest neighbors of the inputs. 
One specification for the nonconformal scores is 
\begin{equation}
    s_i = \Big| \frac{f(\bx_i) - y_i}{\gamma + \xi_i^k} \Big|, 
     \label{eq:def_si}
\end{equation}
where $\xi_i^k$ is the normalized standard deviation of response values for the $i$-th data point and its $k$ neighbors from the training data set. Formally, 
$\xi_i^k = {v_i^k}/{\text{median}(\{v_j^k : j \in \mathcal{I}_{train}\})}$, 
where $\mathcal{I}_{train}$ is the data index set used for the training regression model $f$, $v_i^k$ is the standard deviation of  response values of neighbors found in the training dataset for data $\bx_i$. A higher $v_i^k$ value means larger variance for a prediction at $\mathbf{x}_i$ and thus leads to wider uncertainty bandwidth at $\mathbf{x}_i$.
The tuning parameter $\gamma \geq 0 $ governs the sensitivity of the measure to changes in $\xi_i^k$. Higher $\gamma$ leads to less sensitive measure; when $\gamma = 0$, $s_i = \big| {f(\bx_i) - y_i}/{  \xi_i^k} \big|$.
Instead of normalizing conformal scores using the standard deviation of $f(\mathbf{x}_i)$ for each input $\mathbf{x}_i$, this new measure is normalized using the standard deviation of the response values for its neighbors. 
This approach enables local adaptivity based on local variance instead of requiring a built-in heuristic notion of uncertainty associated with each point.


\begin{algorithm}
\label{vcp}
  \caption{Conformal Prediction}\label{alg:vcp}
  \begin{algorithmic}
    \State{\textbf{Input:}} $\mathcal{D}_{\mathrm{cal}}=\{(\bx_i,y_i)\}, i \in \mathcal{I}_{cal}$, $\bx_{n+1}$, $f:\mathcal{X}\to\mathbb{R}$, $\alpha\in(0,1)$
    \State{\textbf{Output:}} $C_{1-\alpha}(\bx_{n+1})$
    \For{$i \in \mathcal{I}_{cal}$}
      \State $s_i \gets \lvert f(\bx_i) - y_i \rvert$
    \EndFor
    \State $q_{1-\alpha} \gets \text{$\lceil (1-\alpha)(n_2+1)\rceil$-th smallest of } \{s_1,\dots,s_{n_2}\}$
    \State \Return $C_{1-\alpha}(\bx_{n+1}) \gets \big[f(\bx_{n+1})-q_{1-\alpha},\, f(\bx_{n+1})+q_{1-\alpha}\big]$
  \end{algorithmic}
\end{algorithm}

\subsection{Gaussian Process with Automatic Relevance Determination (ARD) Kernel}  \label{sec:gp}

\textbf{Gaussian Processes (GPs)} provide a flexible Bayesian nonparametric framework for regression and classification, where the response variable is modeled as a distribution over functions. Formally, a GP is defined as a collection of random variables:
\[
f(\mathbf{x}) \sim \mathcal{GP}(m(\mathbf{x}), k(\mathbf{x}, \mathbf{x}')),
\]
where $m(\mathbf{x}) = \mathbb{E}[f(\mathbf{x})]$ is the mean function and $\quad k(\mathbf{x}, \mathbf{x'}) = \operatorname{Cov}(f(\mathbf{x}), f(\mathbf{x'}))$ is the covariance kernel. 
Any finite subset $\mathbf{X} = \{\mathbf{x_1}, \mathbf{x_2}, \ldots, \mathbf{x}_n\}$ follows a multivariate normal distribution:

$$
\begin{bmatrix}
f(\mathbf{x}_1)\\
f(\mathbf{x}_2)\\
\vdots\\
f(\mathbf{x}_n)
\end{bmatrix}
\sim \mathcal{N} \left(
\begin{bmatrix}
m(\mathbf{x}_1) \\
\vdots \\
m(\mathbf{x}_n)
\end{bmatrix}
,
\begin{bmatrix}
k(\mathbf{x}_1, \mathbf{x}_1) & k(\mathbf{x}_1, \mathbf{x}_2) & \cdots & k(\mathbf{x}_1, \mathbf{x}_n) \\
k(\mathbf{x}_2, \mathbf{x}_1) & k(\mathbf{x}_2, \mathbf{x}_2) & \cdots & k(\mathbf{x}_2, \mathbf{x}_n) \\
\vdots & \vdots & \ddots & \vdots \\
k(\mathbf{x}_n, \mathbf{x}_1) & k(\mathbf{x}_n, \mathbf{x}_2) & \cdots & k(\mathbf{x}_n, \mathbf{x}_n)
\end{bmatrix}
\right)
$$

One powerful application of GPs is estimating the relevance of each input variable to the response through the \textit{Automatic Relevance Determination} (ARD) kernel \citep{rasmussen2006gaussian}. The ARD kernel assigns a separate length-scale parameter \(\ell_j\) to each input dimension \(x_j\), 
$k_{\boldsymbol\theta}(\mathbf{x}, \mathbf{x}') = k_{\boldsymbol\theta^{'}}(\mathbf{x} \circ \boldsymbol{\ell}^{-1}, \mathbf{x}' \circ \boldsymbol{\ell}^{-1}), 
    \boldsymbol{\ell} = (\ell_1, \dots, \ell_d)$, $\mathbf{x} \circ \boldsymbol{\ell}^{-1} = (\frac{x_1}{\ell_1}, \frac{x_2}{\ell_2}, \cdots, \frac{x_d}{\ell_d})$, $k_{\boldsymbol\theta}$ is a kernel with parameters represented as $\boldsymbol\theta$ including length-scales $\boldsymbol{\ell}$.

Here, a small \(\ell_j\) indicates that the response varies sharply along the \(j\)-th input dimension, implying high relevance of feature \(x_j\), while a large \(\ell_j\) smooths out variation, suggesting that \(x_j\) has little effect. By optimizing the marginal likelihood of the GP with respect to the kernel hyperparameters \(\{\ell_j\}\), one can automatically infer the importance of each input, effectively adjusting for feature importance in a probabilistic and interpretable manner. 

The marginal likelihood of a GP regression model, given training inputs \(X = \{\mathbf{x}_i\}_{i=1}^n\) and outputs \(\mathbf{y}\), is
\[
p(\mathbf{y} \mid X, \boldsymbol{\theta}) = \mathcal{N}\!\big(\mathbf{y} \,\big|\, {m}(\mathbf{x}), K_{\boldsymbol{\theta}} + \sigma_n^2 I \big),
\]
where \(K_{\boldsymbol{\theta}}\) is the covariance matrix computed from the kernel (here the ARD kernel), $K_{\boldsymbol\theta, i,j} = k_{\boldsymbol\theta}(\mathbf{x}_i, \mathbf{x}_j)$, ${m}(\mathbf{x})$ is the mean of $y$ at $\bx$ ,  \(\sigma_n^2\) is the noise variance, and \(\boldsymbol{\theta} = \{\ell_1, \dots, \ell_d, \sigma_n\}\) are kernel hyperparameters. The log marginal likelihood is
\begin{equation}
    \log p(\mathbf{y} \mid X, \boldsymbol{\theta}) = -\tfrac{1}{2}\mathbf{y}^\top (K_{\boldsymbol{\theta}} + \sigma_n^2 I)^{-1}\mathbf{y} - \tfrac{1}{2}\log \big| K_{\boldsymbol{\theta}} + \sigma_n^2 I \big| - \tfrac{n}{2}\log(2\pi).
    \label{eq:gp_lik}
\end{equation}
The length scales $\mathbf{\ell}$ were estimated by maximizing the GP log marginal likelihood with respect to the hyperparameters.
If the \(j\)-th input is irrelevant, the optimizer tends to learn a large $\ell_j$, making the kernel nearly insensitive to changes along that dimension. Conversely, if \(x_j\) strongly influences \(\mathbf{y}\), the optimizer finds a small \(\ell_j\), allowing the GP to vary rapidly along that input. In this way, the ARD mechanism embedded in the GP provides a principled Bayesian method for feature relevance estimation, balancing model fit and complexity through the log marginal likelihood.

Estimating the full GP is computationally expensive for high-dimensional data, with computational complexity scaling as $\mathcal{O}(n^3)$, where $n$ denotes the number of training observations in our setting. Several scalable GP approximation methods have been developed, including inducing point methods \citep{snelson2006sparse}, covariance tapering \citep{furrer2006covariance}, and the Vecchia approximation \citep{vecchia1988estimation,Katzfuss2021}. At a high level, the Vecchia method approximates the joint Gaussian density using conditional distributions conditioned on subsets of neighboring observations.
In our framework, we adopt the Vecchia approximation because it preserves local dependence structures through nearest-neighbor conditioning, which aligns naturally with our overall approach of leveraging neighbor information for uncertainty quantification (UQ). This method is implemented in the \texttt{GpGp} package developed by \cite{guinness2019gpgp}.

%% file: method.tex
\subsection{Adaptive Conformal Prediction with Gaussian Process Regression for Image Based Models }

While adaptive conformal prediction using nearest neighbors guarantees a desired coverage rate, its application to image-based models is challenging. The primary difficulty lies in defining a meaningful set of ``nearest neighbors" for images. Standard image similarity metrics, such as the Structural Similarity Index (SSIM) and Peak Signal-to-Noise Ratio (PSNR), are inadequate as they only assess pixel-level variations and ignore the relationship between an image and the response variable of interest. To address this, we propose a novel framework that integrates the computational benefits of nearest-neighbor adaptive conformal prediction with the feature-relevance estimation capabilities of GPs. This approach yields prediction sets with guaranteed coverage for image-based regression or classification models.

\textbf{Problem Formulation:}
Formally, given datasets $\mathcal{D}_{train}$, $\mathcal{D}_{cal}$ as defined in section \ref{sec:background_cp}, and set of new data $\mathcal{D}_{test} = \{(\bx_i, y_i)\}$. Let $X_{train} = \{\bx_i, i \in \mathcal{I}_{train}\} \in \mathbb{R}^{n_1 \times d_1 \times d_2 \times d_3}$ be a set of $n_1$ images and $\mathbf{y}_{train} =  \{y_i, i \in \mathcal{I}_{train}\}\in \mathbb{R}^{n_1}$ be the corresponding responses. Each image has $d_1$ channels, a height of $d_2$ and a width of $d_3$, $d = d_1 \times d_2 \times d_3$. Given a model ${f}$ trained on this data, our goal is to generate an adaptive prediction set, $\mathcal{C}_{1-\alpha}(\mathbf{x}_i)$, for any new test image $\mathbf{x}_i \in X_{test}$, such that the marginal coverage probability for the response at $\mathbf{x}_i$ meets a user-defined level, $1-\alpha$
$$
\mathbb{P}\big( y_{i} \in \mathcal{C}_{1-\alpha}(\mathbf{x}_i) \big) \geq 1-\alpha.
$$
The performance of our method is evaluated based on the adaptivity of the prediction set sizes and the empirical coverage rate on the test data.

\textbf{Proposed Methodology:}
Our approach consists of three main stages: dimensionality reduction, feature relevance estimation, and nearest-neighbor-based conformalization.

First, we project images into a lower-dimensional latent space of size $m$, where $m \ll d$. This is based on the manifold hypothesis (\cite{tenenbaum2000global}), which argues that the essential information in high-dimensional data can be captured in a much lower-dimensional representation. In this paper we use Principal Component Analysis (PCA) for this purpose. Other embeddings like the latent space from an autoencoder could also be used. We denote the projection by $h: \mathbb{R}^d \to \mathbb{R}^m$ and the condensed data by $\tilde{X} = h(X)$.

Second, we estimate the relevance of each latent feature with respect to the response variable, $\mathbf{y}$. This is achieved by modeling the relationship between the low-dimensional training data and its responses using a Gaussian Process with an Automatic Relevance Determination (ARD) kernel. The ARD kernel yields a unique length-scale parameter, $\ell_j$, for each latent dimension $j$, where smaller length-scales indicate higher relevance to the prediction task. In the case when data size is large ($> 10^4$), we use scalable Vecchia Gaussian Process (\cite{Katzfuss2021}), which imposes sparsity in the covariance matrix by assuming conditionally independent of distant data given a small set of its neighbors. 

Third, for each test sample $\mathbf{x}$, we identify its $k$ nearest neighbors from the training set. The search is performed using a scaled Euclidean distance metric weighted by the inverse of the learned ARD length-scales. This ensures that dimensions more relevant to the response variable have a greater influence on the distance calculation. The set of $k$ nearest neighbors, $N_k(\mathbf{x})$, is defined as:
\begin{equation}
    N_k(\mathbf{x}) = \operatorname*{arg\,min}_{S \subseteq \mathcal{I}_{train}=\{1,\dots,n\},\, |S|=k} \sum_{i \in S} D(\mathbf{x}, \mathbf{x}_i)
    \label{eq:NN}
\end{equation}
where the distance metric $D(\cdot, \cdot)$ is:
$$
D(\mathbf{x}, \mathbf{z}) = \sqrt{\sum_{j=1}^{d}\frac{(x_j - z_j)^2}{\ell_j^2}}
$$
To perform this search efficiently, we utilize a KD-tree algorithm, which has an average-case query time complexity of $O(k \log n)$. This neighborhood set, $N_k(\mathbf{x})$, is then used to calculate a local conformity score using Equation \ref{eq:def_si}, which in turn determines the bandwidth of the final prediction interval. The remaining steps follow the standard adaptive conformal prediction procedure. Our complete framework is summarized in Algorithm~\ref{alg:aCP}.

\begin{algorithm}[H]
\caption{Adaptive Conformal Prediction using GP-Informed Nearest Neighbors}
\label{alg:aCP}
\begin{algorithmic}
\Statex \textbf{Input:} $X_{\text{train}}, \mathbf{y}_{\text{train}}$,  $X_{\text{cal}}, \mathbf{y}_{\text{cal}}$,  $\mathbf{x}_{\text{test}}$, ${f}$
\Statex \textbf{Parameters:}  $k$,  $d$, $\gamma$,  $\alpha$
\Statex \textbf{Output:} $\mathcal{C}_{1-\alpha}(\mathbf{x}_{\text{test}})$.

\State Dimension reduction: $[\tilde{X}_{train}, \tilde{X}_{cal}, \tilde{\bx}_{test}] \leftarrow h([{X}_{train}, {X}_{cal}, {\bx}_{test}])$.
\State Estimate $\boldsymbol{\ell}$ via maximize equation \eqref{eq:gp_lik} with ARD kernel $\kappa$.%
\For{ $\forall \tilde{\bx_i} \in \{\tilde{X}_{train}, \tilde{X}_{cal}, \tilde{\bx}_{test}\}$}
    \State Find  $N_k(\tilde{\mathbf{x_i}})$ via equation \eqref{eq:NN}
    \State $v_i \leftarrow \text{std}(\{y_j\}_{j \in N_k(\tilde{\mathbf{x}}_i)})$.
\EndFor
\State $v_{\text{median}} \leftarrow \text{median}(\{v_i\}_{i \in \mathcal{I}_{train}})$.
\For{$\forall \tilde{\bx_i} \in  \tilde{X}_{cal}$}
    \State $\xi_i \leftarrow v_i / v_{\text{median}}$.
    \State $s_i \leftarrow |y_i - {f}(\mathbf{x}_i)| / (\gamma + \xi_i)$.
\EndFor

\State Compute quantile: ${q} = \text{Quantile}(\{s_i\}, 1-\alpha)$.
\State \Return  $\mathcal{C}(\mathbf{x}_{\text{test}}) = \left[ {f}(\mathbf{x}_{\text{test}}) \pm {q} \cdot (\gamma +\xi_{\text{test}}) \right]$.

\end{algorithmic}
\end{algorithm}

%% file: application.tex
\section{Motivating Application: Inertial Confinment Fusion} \label{sec:application}
\subsection{Setups}
\begin{figure}
	\centering
	\includegraphics[trim={0cm 3cm 0cm 3cm},clip, width =1\linewidth]{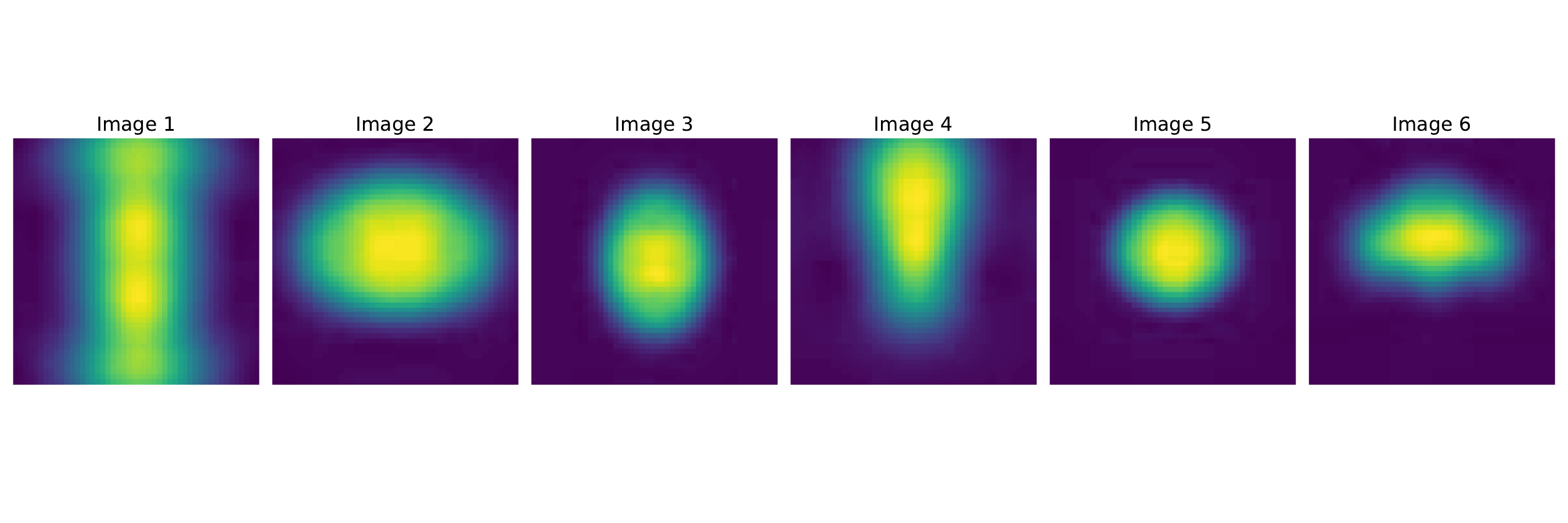}
    \caption{\textbf{Examples of X-ray images from the BigFoot dataset.} From left to right, the images were selected based on their relatively large scores along principal components 1–6, respectively (approximately the 500th largest score among 90,000 images for each component). These images highlight characteristic image features associated with each principal component.}
    \label{fig:real6}
\end{figure}

Our work focuses on the multi-modal diffusion model introduced by Jones et al. \cite{jones2025}, which is designed to model the joint and conditional distributions of scalar and image data generated in inertial confinement fusion experiments. Specifically, this joint multi-modal diffusion-based generative model acts as a forward and inverse surrogate for the radiation hydrodynamic code HYDRA (\cite{{Marinak_2001}}), enabling prediction of simulation inputs, our response variable $y$, from their corresponding radiation hydrodynamic images, $\mathbf{x}$. Sample images are visualized in Figure \ref{fig:real6}. 

HYDRA radiation-hydrodynamics simulations played a central role in the “Bigfoot” campaign by providing a computational means to evaluate capsule and drive design parameters (\cite{Nora2017}). In parallel with experiments performed at the National Ignition Facility (NIF) from 2016 to 2017, which sought to increase implosion velocity and reduce low-mode asymmetries by varying laser drive and capsule scale, a large ensemble of 2D HYDRA simulations was generated (\cite{Casey2018}). This ensemble was constructed by varying nine capsule inputs and produced approximately 90,000 high-resolution simulations, along with a suite of synthetic scalar outputs and diagnostic images. Producing these data required more than 20 million CPU-hours, making the Bigfoot simulation ensemble one of the largest ICF simulation datasets containing X-ray images.

The goal here is to augment the pre-trained point prediction model, ${f}$, with rigorous uncertainty quantification. Because $f$ is a diffusion model, a heuristic notion of uncertainty is available by taking the standard deviation of $n$ draws from the model. This is not true for many ML models that are deterministic and we aim to develop a method that can be applied more generally. The diffusion model variability provides a reasonable comparison. We aim to construct {adaptive prediction intervals}, $C_{1-\alpha}(\mathbf{x}_i)$, for any given image $\mathbf{x}_i \in \mathbb{R}^{48 \times 48}$, such that the true input $y_i$ is contained within the interval with a user-specified probability, formally $\mathbb{P}\big( y_{i} \in C_{1-\alpha}(\mathbf{x}_i) \big) \geq 1-\alpha$.

\textbf{Data and Model Setup:}
The existing model ${f}$ was trained on a large dataset of 90,000 image-input pairs simulated by HYDRA (\cite{Marinak_2001}), which we denote as $\mathcal{D}_{\text{train}}$. For our uncertainty analysis, we utilize an additional 2,630 held-out samples. We randomly partition this unused data evenly into a {calibration set} $\mathcal{D}_{\text{cal}}$ and a {test set} $\mathcal{D}_{\text{test}}$, each containing 1,315 data points. This random split ensures that the calibration and test data are drawn from the same distribution, which is a key assumption for conformal prediction. The model generatively predicts capsule simulation inputs $\mathbb{R}^9$ (response variables) from simulation output X-ray images $\mathbb{R}^{48 \times 48}$. While X-ray images can be measured in experiment, capsule inputs are challenging to measure and therefore must be inferred from simulation data. These inputs include symmetry parameters and laser energy delivery, which provide crucial information about possible failure modes in ICF experiments. 

In this work, we focus our analysis on predicting two inputs from this ensemble in particular — the drive mode 1 and mode 2 asymmetries. In indirect drive ICF, lasers do not strike the capsule directly but rather heat the surface of an outer container called a ``hohlraum” which produces an X-ray radiation bath that subsequently drives the capsule implosion. Under ideal conditions, this radiation profile is uniform and the capsule remains round as it compresses. More commonly, however, there are small perturbations to the drive which are amplified during the implosion and degrade performance. Mode 1 refers to a net perturbation in any direction, and mode 2 refers to more or less energy being delivered to the pole or equator of the capsule. X-ray images are critical for diagnosing these drive asymmetries, however the mapping is complex and degenerate, as multiple types of perturbations can produce similar images.

\begin{figure}
\centering
    \includegraphics[trim={0cm 3cm 0cm 2cm},clip, width =1\linewidth]{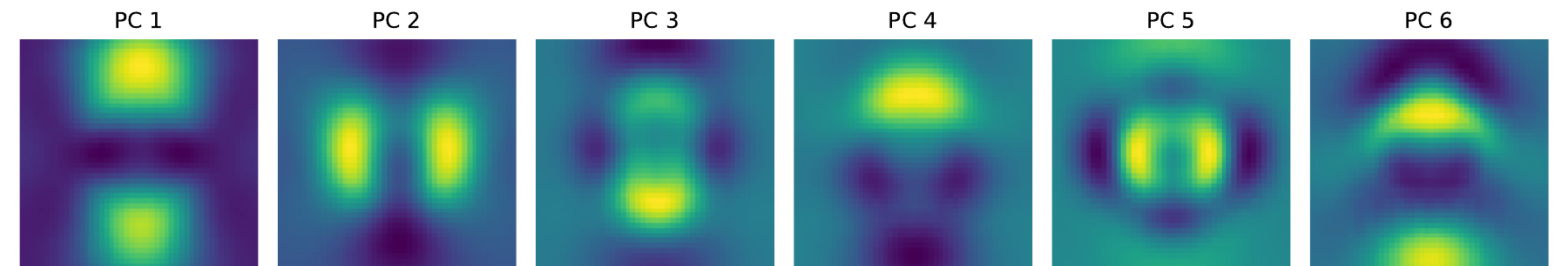}
    \caption{\textbf{Visualization of the first six principal components (PCs)} obtained from PCA of the 90,000 training X-ray images in the BigFoot dataset.Each component is displayed as an eigenimage by reshaping the corresponding PCA loading vector to the original image dimensions. The PCs represent the dominant modes of variation in the training data, with PC1 explaining the greatest amount of variance and subsequent PCs explaining progressively less.
}
    \label{fig:6pc}
\end{figure}

\textbf{Dimension Reduction:}
We first reduce the dimensionality of the input images from $48 \times 48$ to 6 using {Principal Component Analysis (PCA)}. These 6 components capture 90\% of the data variance. The visualization of the top 6 principal components (PC) are shown in Figure \ref{fig:6pc}.  This projection captures the most significant features of the X-ray images while drastically reducing computational cost for late steps. For example, PC2 can be deemed associated with response variable mode 2 asymmetries. Also, PCA is more computationally efficient compared to other techniques like using Variational Autoencoder(VAE), which can suffer from its sensitivity to the architecture of the neural network and hyperparameter tuning. 

\textbf{Relevance Estimation:} To learn the relative importance of the input features with respect to the response variables of interest, we modeled the relationship between the latent-space inputs and the response variables using a Gaussian process with an ARD Mat'ern kernel (smoothness parameter fixed at 1.5) implemented via the \texttt{GpGp} package. The estimated relevance for PC 5 was lower than that of the other principal components for both response variables, indicating that PC 5 is relatively less informative for the prediction task and contributes less to the construction of the neighborhood-based conditional sets in later stages.

\textbf{Uncertainty Estimation:}
Our adaptive uncertainty estimate, $\xi_i$, is derived using a nearest-neighbors approach in the reduced 6-dimensional PCA space. For each point in the calibration and test sets, we use a KD-tree to efficiently find its $k = 30$ nearest neighbors within the large training set $\mathcal{D}_{\text{train}}$ under the scaled Euclidean distance metric. We chose $k=30$ because we observed that pairwise distance stabilized and become insignificant after the $31$-th neighbor in our empirical study. The variance of the true response variables ($y$ values) corresponding to these $k$ neighbors is then calculated and scaled to serve as our local uncertainty metric, $\xi_i$.

\textbf{Conformalization:}
We use a split conformal prediction scheme to generate the intervals. We first define a {non-conformity score} for each sample $(\mathbf{x}_i, y_i)$ in the calibration set, which measures the normalized error of the model:
    $s_i = {|{f}(\mathbf{x}_i) - y_i|}/{\xi_i}.$
Next, we calculate the quantile ${q}_{1-\alpha}$ of the non-conformity score. Finally, the adaptive prediction interval for a new test image $\mathbf{x}_j \in \mathcal{D}_{\text{test}}$ is constructed as:
$C_{1-\alpha}(\mathbf{x}_{j}) = \big[ {f}(\mathbf{x}_{j}) \pm {q}_{1-\alpha} \cdot \xi_j \big].$
This formulation ensures that the prediction intervals are wider for test points with higher estimated local variance ($\xi_j$) and narrower for those in well-sampled, low-variance regions, all while maintaining the guaranteed long-run coverage of $1-\alpha$.

\subsection{Results}

\begin{figure}[htbp]
  \centering

  \newcommand{\panelwidth}{0.88\linewidth}
  \newcommand{\imgwidth}{0.315\linewidth}

  \begin{tikzpicture}[every node/.style={inner sep=0,outer sep=0}]
    \node (panel) {
      \begin{minipage}{\panelwidth}
        \centering
        \setlength{\tabcolsep}{1.5pt}
        \renewcommand{\arraystretch}{0}

        \begin{tabular}{@{}ccc@{}}
          \includegraphics[trim=6 6 6 6, clip, width=\imgwidth]{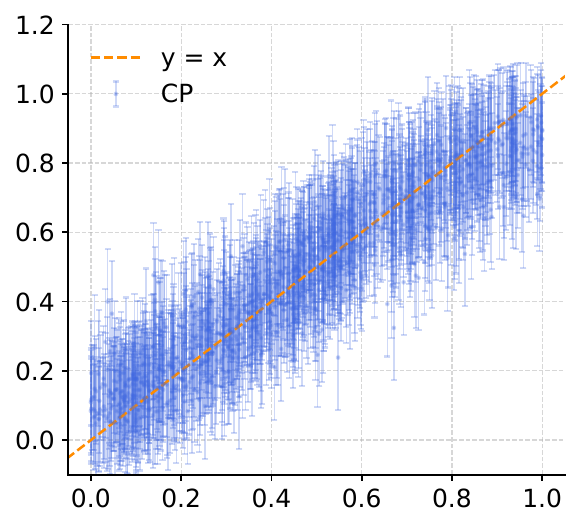} &
          \includegraphics[trim=6 6 6 6, clip, width=\imgwidth]{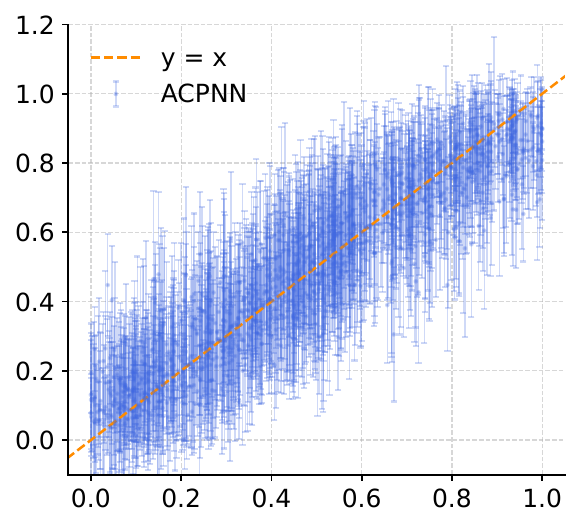} &
          \includegraphics[trim=6 6 6 6, clip, width=\imgwidth]{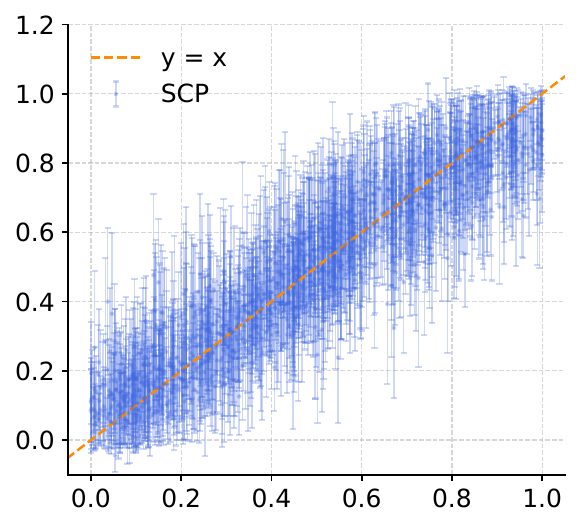} \\
          \includegraphics[trim=6 6 6 6, clip, width=\imgwidth]{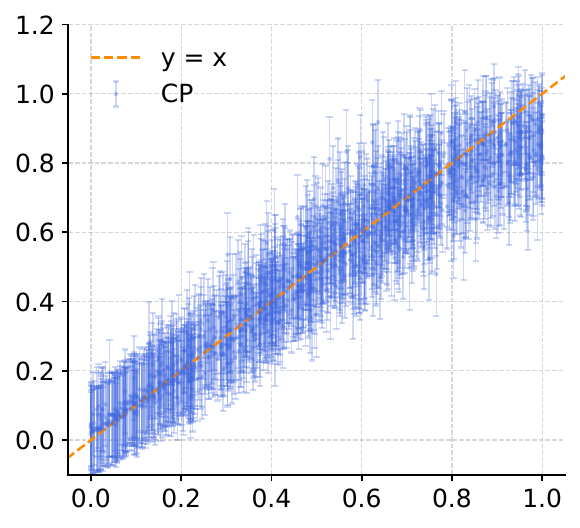} &
          \includegraphics[trim=6 6 6 6, clip, width=\imgwidth]{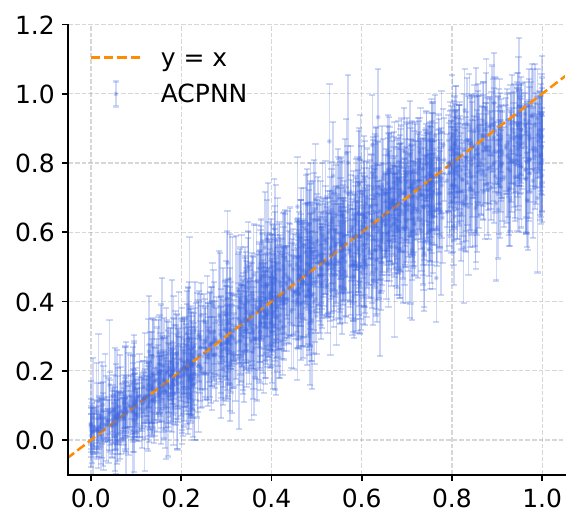} &
          \includegraphics[trim=6 6 6 6, clip, width=\imgwidth]{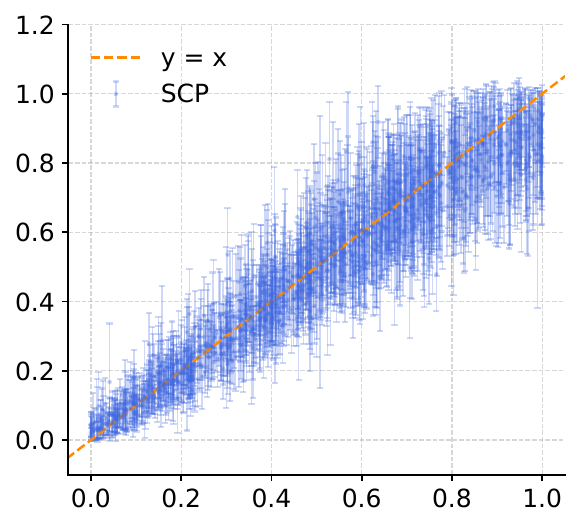}
        \end{tabular}
      \end{minipage}
    };

    \node[rotate=90, align=center]
      at ($ (panel.north west)!0.25!(panel.south west) + (1mm,0) $)
      {\tiny mode 2 asymmetry};

    \node[rotate=90, align=center]
      at ($ (panel.north west)!0.75!(panel.south west) + (1mm,0) $)
      {\tiny mode 1 asymmetry};

    \node[above=1mm]
      at ($ (panel.north west)!0.1666!(panel.north east) $)
      {\footnotesize CP};

    \node[above=1mm]
      at ($ (panel.north west)!0.5000!(panel.north east) $)
      {\footnotesize ACPNN};

    \node[above=1mm]
      at ($ (panel.north west)!0.8333!(panel.north east) $)
      {\footnotesize SCP};

    \node[below=1mm of panel.south]
      {\normalsize True Values};

    \node[rotate=90, left=2mm of panel.west, anchor=south]
      {\normalsize Predicted Values};

  \end{tikzpicture}

  \caption{Visualization comparing conformal prediction sets with the true response values across different conformal prediction methods. The first row displays results for the predicted values of mode 2 asymmetry while the second row shows predictions for mode 1 asymmetry. All three methods achieve the desired coverage of the true responses. The compactness of the conformal sets reflects model performance, with adaptive conformal prediction and scaled conformal prediction producing similarly tight prediction intervals.}
  \label{fig:y_sets}
\end{figure}

We benchmark our method against conformal prediction (CP) and scaled conformal prediction (SCP) by evaluating predictive accuracy, uncertainty bandwidths, and input-conditional coverage rates. Our results show that all three methods achieve the desired coverage levels, successfully capturing the true response values. Moreover, both SCP and ACPNN produce more input-adaptive prediction intervals than CP (see Figure~\ref{fig:y_sets}). We also investigated the prediction performance of ACPNN using different numbers of ensembles, with the corresponding results presented in Appendix~\ref{y_sets_acpnn} (Figure~\ref{fig:y_sets_acpnn}).

\begin{figure}[htbp]
    \centering

    \begin{subfigure}[t]{0.9\textwidth}
        \centering
        \includegraphics[width=\linewidth]{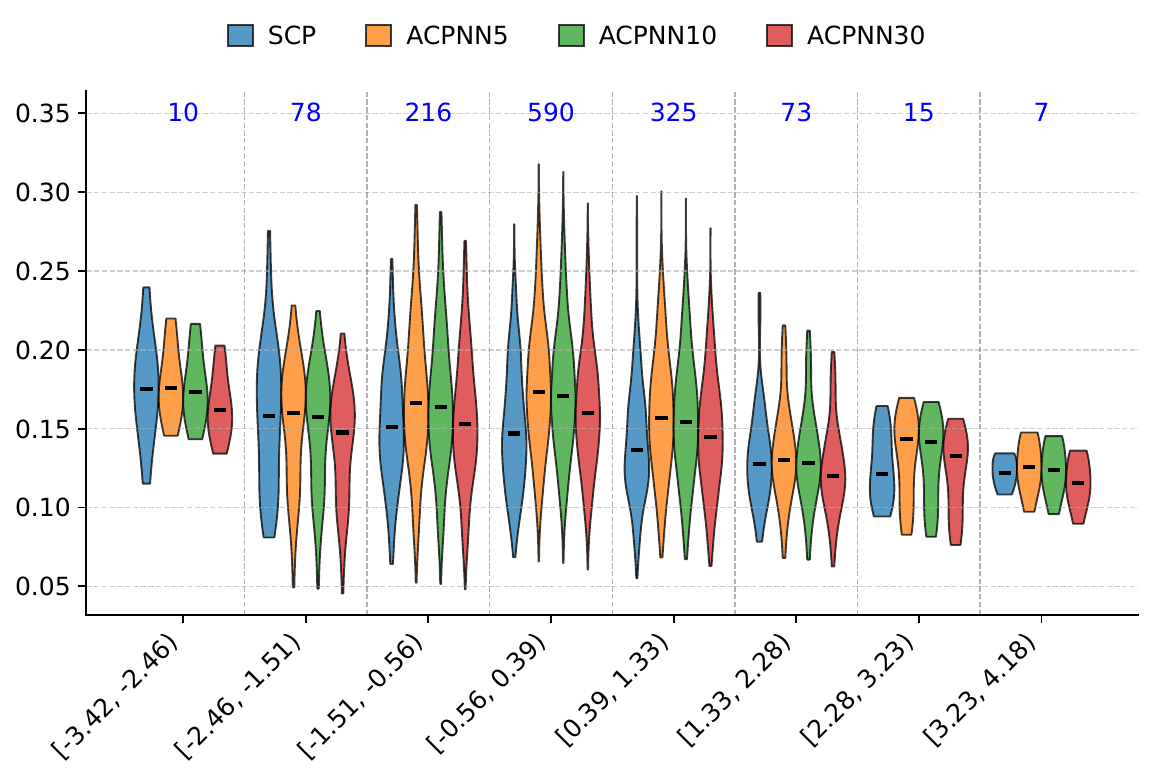}
        \caption{}
        \label{fig:j3pc6violin}
    \end{subfigure}
    \hfill
    \begin{subfigure}[t]{0.9\textwidth}
        \centering
        \includegraphics[width=\linewidth]{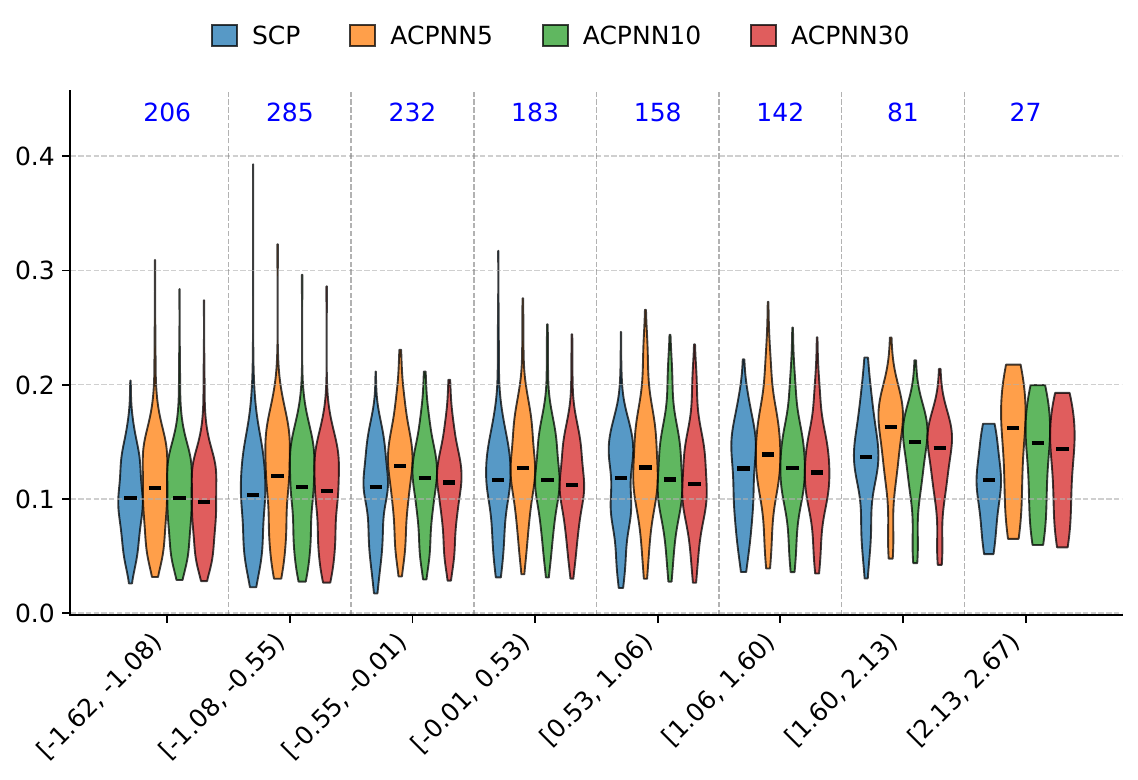}
        \caption{}
        \label{fig:j1pc1violin}
    \end{subfigure}

    \caption{Violin plots of uncertainty magnitudes across input bins. In Panel (a) PC 6 is used as the binned input and mode 2 asymmetry as the response variable; in panel (b) PC 1 is used as the input and mode 1 asymmetry as the response variable. The x-axis represents the input regions (bins), and the y-axis denotes the magnitude of predictive uncertainty. Numbers shown above each panel indicate the number of test samples within each corresponding input region.
}
    \label{fig:violin_uq_bin}
\end{figure}

We further examine the conformal interval widths of ACPNN with different ensemble sizes (3, 5, 10, and 30) and compare them with SCP, which is based on 30 ensembles throughout this paper. A summary of the results is provided in Table~\ref{tab:results}, with representative examples visualized in Figure~\ref{fig:violin_uq_bin}. 

\begin{table}[htbp]
\centering
\resizebox{\textwidth}{!}{
\begin{tabular}{lcccccccc}
\toprule
& \multicolumn{4}{c}{Mode 2 asymmetry} & \multicolumn{4}{c}{Mode 1 asymmetry} \\
\cmidrule(lr){2-5} \cmidrule(lr){6-9}
Method & Q1 & Median & Q3 & Cov. Rate & Q1 & Median & Q3 & Cov. Rate \\
\midrule
CP       & 0.152 & 0.152 & 0.152 & 0.898 & 0.121 & 0.121 & 0.121 & 0.898 \\
SCP      & 0.119 & 0.144 & 0.174 & 0.905 & 0.079 & 0.112 & 0.141 & 0.890 \\
ACPNN3   & 0.149 & 0.179 & 0.211 & 0.900 & 0.105 & 0.143 & 0.177 & 0.902 \\
ACPNN5   & 0.137 & 0.165 & 0.195 & 0.898 & 0.094 & 0.128 & 0.158 & 0.899 \\
ACPNN10  & 0.135 & 0.162 & 0.192 & 0.903 & 0.086 & 0.117 & 0.145 & 0.884 \\
ACPNN30  & 0.126 & 0.152 & 0.179 & 0.884 & 0.084 & 0.113 & 0.140 & 0.886 \\
\bottomrule
\end{tabular}
}
\caption{Comparison of uncertainty interval distributions (Q1, median, Q3) and empirical coverage rates across methods for two drive asymmetry response variables: mode 2 asymmetry and mode 1 asymmetry.}
\label{tab:results}
\end{table}

Our experiments indicate that when the underlying diffusion model demonstrates varying levels of predictive performance (e.g., it predicts mode 1 asymmetry more accurately than mode 2 asymmetry), ACPNN achieves interval widths comparable to those of SCP, particularly in input regions with abundant data (see Figure~\ref{fig:j1pc1violin}).

In contrast, when predictive performance is weaker, SCP tends to perform better in data-rich regions, whereas ACPNN shows advantages in data-scarce regions (see Figure~\ref{fig:j3pc6violin}). 
This observation can be explained by the fact that interval widths under SCP are more closely tied to model performance, as ensemble outputs are heavily used for both mean and variance estimation. In contrast, ACPNN leverages information from nearest neighbors in the training data to estimate variance and is therefore less dependent on model performance. This feature of ACPNN creates opportunities for uncertainty quantification in scenarios where data are limited and the model is not well trained.
We also observe that increasing the number of ensembles in ACPNN consistently reduces uncertainty bandwidths in both scenarios. This effect stabilizes beyond approximately 10 ensembles when the model exhibits strong predictive performance (see Figure~\ref{fig:j1pc1violin}).
Overall, both ACPNN and SCP maintain stable coverage rates across different input regions. While the two methods produce comparable uncertainty bandwidths on average, ACPNN tends to yield slightly narrower prediction intervals in regions with limited data. This suggests that, under data-scarce or extreme conditions, ACPNN provides more compact and informative uncertainty quantification, highlighting its advantage when training data are limited.

%% file: conclusion.tex
\section{Conclusion and Discussion}

Our experimental results demonstrate the effectiveness of the proposed method (ACPNN) in its application to ICF data. The method achieves the desired empirical coverage while providing adaptive, image-dependent conformal prediction intervals for code outputs, enabling more nuanced uncertainty quantification.

While scaled conformal prediction (SCP) retains some advantage in certain regimes—yielding narrower intervals in regions with abundant data even when model performance is weak—its applicability is limited in more constrained settings. In particular, when generating multiple ensembles is computationally prohibitive, as is often the case when emulator evaluations are expensive, ACPNN provides a clear advantage and is better suited for practical deployment.

Moreover, in scenarios where ensemble generation is infeasible due to the deterministic nature of the model (e.g., the lane-keeping CNN introduced in \cite{CNNangleChen2017}, the Fourier Neural Operator in \cite{li2021fourier}, and vision transformer–style architectures such as ConvNeXt in \cite{liu2022convnet}), SCP is usually not applicable, whereas ACPNN remains effective and delivers competitive performance even with a single ensemble. These advantages are particularly relevant in scientific settings, where data is often limited and computational constraints favor lightweight models over more expensive approaches such as diffusion models, further reinforcing the practicality of ACPNN.

Finally, although our framework employs a simple dimensionality reduction technique (PCA), it still achieves competitive overall performance, indicating robustness and suggesting potential gains from incorporating more advanced representation learning methods.

Looking ahead, a compelling direction for future work is to extend this method to scenarios involving distribution shifts, i.e when training data and testing data have different distributions, as if often the case when comparing ICF simulations to experiments. A key challenge in this context will be to develop a meaningful and robust definition of \textquotedblleft nearest neighbor", and defining weighted conformal scores and weighted quintile, especially in the case when distributions of images are more complicated like bimodal distribution. This direction potentially involve exploring density ratio estimation in a learned latent space. Further research could also investigate the theoretical properties of the proposed adaptive sets and explore applications in other high-stakes domains where calibrated uncertainty is critical.

%% file: appendix.tex
\section{Additional Simulation Results}
\label{y_sets_acpnn}

\begin{figure}[htbp]
  \centering
  
  \newcommand{\panelwidth}{0.88\linewidth}
  \newcommand{\imgwidth}{0.315\linewidth}
  
  \begin{tikzpicture}[every node/.style={inner sep=0,outer sep=0}]
    \node (panel) {
      \begin{minipage}{\panelwidth}
        \centering
        \setlength{\tabcolsep}{1.5pt}
        \renewcommand{\arraystretch}{0}
        \begin{tabular}{ccc}
          \includegraphics[trim=6 6 6 6, clip, width=\imgwidth]{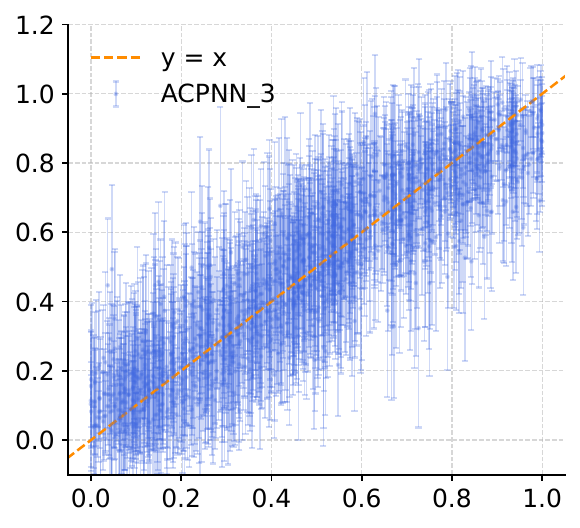} &
          \includegraphics[trim=6 6 6 6, clip, width=\imgwidth]{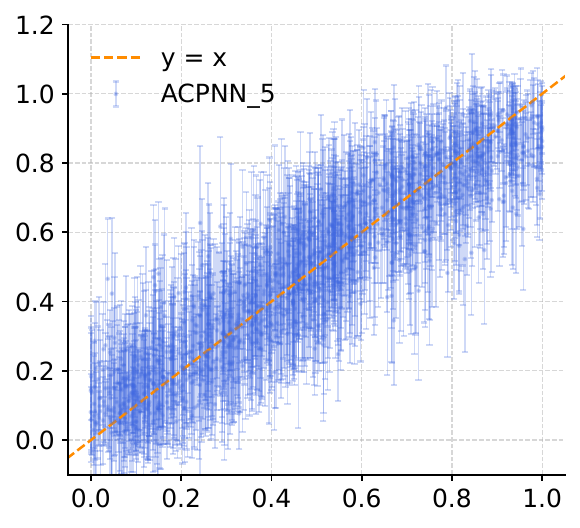} &
          \includegraphics[trim=6 6 6 6, clip, width=\imgwidth]{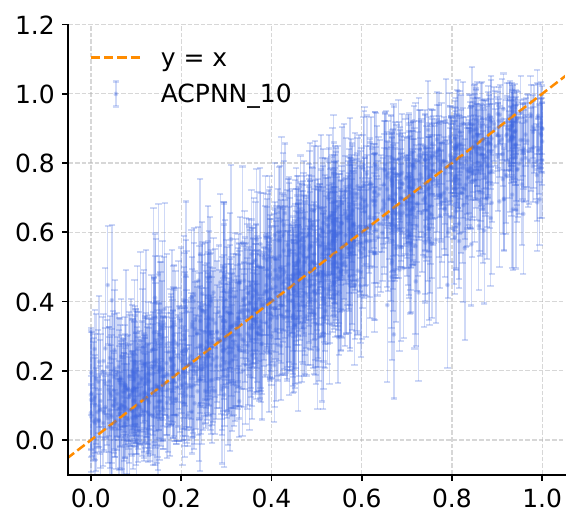} \\
          \includegraphics[trim=6 6 6 6, clip, width=\imgwidth]{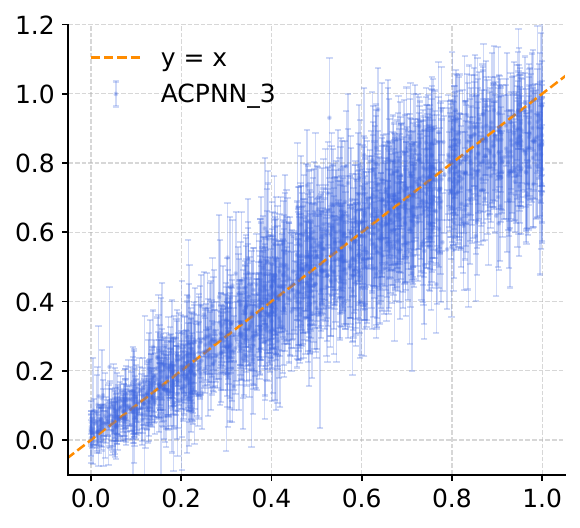} &
          \includegraphics[trim=6 6 6 6, clip, width=\imgwidth]{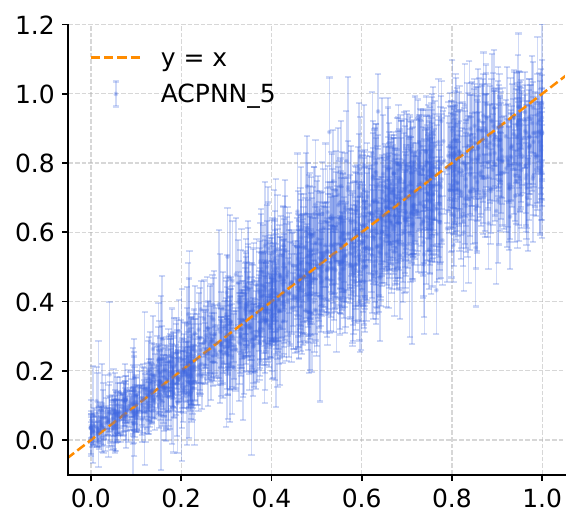} &
          \includegraphics[trim=6 6 6 6, clip, width=\imgwidth]{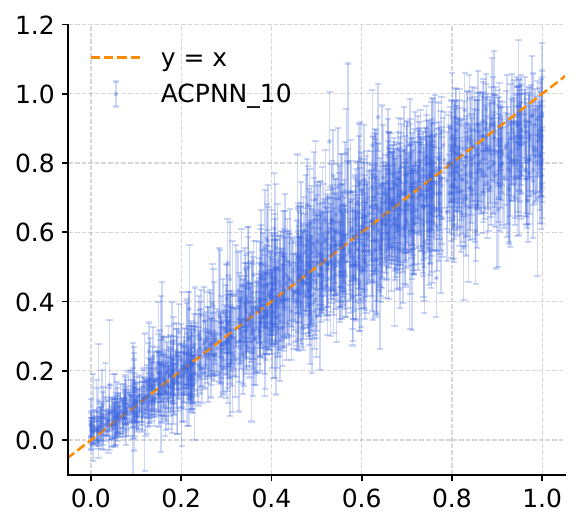}
        \end{tabular}
      \end{minipage}
    };

    \node[rotate=90, align=center]
      at ($ (panel.north west)!0.25!(panel.south west) + (1mm,0) $)
      {\tiny Mode 2 Asymmetry};

    \node[rotate=90, align=center]
      at ($ (panel.north west)!0.75!(panel.south west) + (1mm,0) $)
      {\tiny Mode 1 Asymmetry};

    \node at ($ (panel.south west)!0.1666!(panel.south east) + (0,85mm) $)
      {\footnotesize ACPNN3};
    \node at ($ (panel.south west)!0.5000!(panel.south east) + (0,85mm) $)
      {\footnotesize ACPNN5};
    \node at ($ (panel.south west)!0.8333!(panel.south east) + (0,85mm) $)
      {\footnotesize ACPNN10};

    \node[below=0mm of panel.south]
      {\normalsize True Values};
    \node[rotate=90, left=3mm of panel.west, anchor=south]
      {\normalsize Predicted Values};

  \end{tikzpicture}
      \caption{Visualization comparing conformal prediction sets with the true response values across ACPNN method with different number of ensembles. The first row displays results for the predicted values of mode 2 asymmetry, while the second row shows predictions for mode 1 asymmetry. All three methods achieve the desired coverage of the true responses. The compactness of the conformal sets reflects method performance, with increase of number of ensembles used in the calculation, higher compactness of the conformal sets.}
    \label{fig:y_sets_acpnn}
\end{figure}

Additional prediction performance results for the ACPNN method using different numbers of ensembles are shown in Figure~\ref{fig:y_sets_acpnn}.

